\documentclass{svproc}
\usepackage{type1cm}        
\usepackage{graphicx}        

\usepackage[bottom]{footmisc}

\usepackage{newtxmath}       

\usepackage{xcolor}

\usepackage{biblatex}
\usepackage[acronym]{glossaries}
\usepackage{glossaries-extra}
\usepackage[super]{nth}
\setabbreviationstyle[acronym]{long-short}
\newacronym{rl}{RL}{Reinforcement Learning}
\newacronym{sb3}{SB3}{Stable Baselines3}
\newacronym{dll}{DLL}{Dynamic Link Library}
\newacronym{dof}{DOF}{Degrees of freedom}
\newacronym{ppo}{PPO}{Proximal Policy Optimization}
\newacronym{mpc}{MPC}{Model Predictive Control}
\newacronym{hil}{HIL}{Hardware-in-the-Loop}

\usepackage{cleveref}
\crefname{section}{Sect.}{Sect.}
\Crefname{section}{Section}{Sections}
\crefname{figure}{Fig.}{Fig.}
\crefname{lstlisting}{Listing}{listings}
\Crefname{lstlisting}{Listing}{Listings}

\usepackage{subcaption}

\usepackage{listings}
\definecolor{codegreen}{rgb}{0,0.6,0}
\definecolor{codegray}{rgb}{0.5,0.5,0.5}
\definecolor{codepurple}{rgb}{0.58,0,0.82}
\definecolor{backcolour}{rgb}{0.95,0.95,0.92}

\lstdefinestyle{mystyle}{
    backgroundcolor=\color{backcolour},   
    commentstyle=\color{codegreen},
    keywordstyle=\color{magenta},
    numberstyle=\tiny\color{codegray},
    stringstyle=\color{codepurple},
    basicstyle=\ttfamily\footnotesize,
    breakatwhitespace=false,         
    breaklines=true,                 
    captionpos=b,                    
    keepspaces=true,                 
    numbers=left,                    
    numbersep=5pt,                  
    showspaces=false,                
    showstringspaces=false,
    showtabs=false,                  
    tabsize=2
}

\usepackage{url}

\usepackage{comment}

\begin{document}
    \mainmatter              
    \title{
        Python-Based Reinforcement Learning on Simulink Models
        \thanks{This preprint, accepted at SMPS24, has not undergone peer review or any post-submission improvements or corrections.}
    }
    \titlerunning{Python-Based RL on Simulink Models}  
    %
    \author{Georg Schäfer\inst{1,2,3} \and Max Schirl\inst{2,3} \and Jakob Rehrl\inst{1,2} \and Stefan Huber\inst{1,2} \and Simon Hirlaender\inst{3}}
    \authorrunning{G. Schäfer et al.} 
    %
    %
    \institute{Josef Ressel Centre for Intelligent and Secure Industrial Automation, Salzburg, Austria  \and Salzburg University of Applied Sciences, Salzburg, Austria \and Paris Lodron University of Salzburg, Salzburg, Austria\\ \email{georg.schaefer@fh-salzburg.ac.at}}
    
    \maketitle              
    
    \begin{abstract}
        This paper proposes a framework for training Reinforcement Learning agents using Python in conjunction with Simulink models. Leveraging Python's superior customization options and popular libraries like Stable Baselines3, we aim to bridge the gap between the established Simulink environment and the flexibility of Python for training bleeding edge agents. Our approach is demonstrated on the Quanser Aero 2, a versatile dual-rotor helicopter. We show that policies trained on Simulink models can be seamlessly transferred to the real system, enabling efficient development and deployment of Reinforcement Learning agents for control tasks. Through systematic integration steps, including C-code generation from Simulink, DLL compilation, and Python interface development, we establish a robust framework for training agents on Simulink models. Experimental results demonstrate the effectiveness of our approach, surpassing previous efforts and highlighting the potential of combining Simulink with Python for Reinforcement Learning research and applications.
        
        \keywords{Reinforcement Learning, Simulink, Python, Stable Baselines3}
    \end{abstract}

    \section{Introduction}

    \gls{rl} has recently gained much interest in the field of control of dynamical systems~\cite{li2017deep, padakandla2021survey}. Often, the design of the \gls{rl} policy is not done directly using the real environment, but instead using a simulation model of the real system that mimics the environment. A common toolset for setting up the simulation model is MATLAB/Simulink. Specifically, the realization of the simulation model by means of block diagrams with the help of Simulink is an effective and well established workflow. However, if developing bleeding edge \gls{rl} agents is the goal, Python may be seen as the de facto standard in the industry. OpenAI Baselines, \gls{sb3}, RLlib, TF-Agents among other sets of implementations are only available in Python and offer superior customization options when it comes to the creation of custom agents and environments. In the present paper, a framework combining these two software environments will be proposed. For that purpose, a lab-scale model of a mechatronic system will be utilized: the Quanser Aero~2. The Canadian manufacturer Quanser established itself as the leading company for the development and manufacturing of advanced control systems and educational platforms for teaching and research in engineering disciplines such as robotics, mechatronics, and control theory.

    \subsection{Motivation}
    Our goal is to train a \gls{rl} policy using Python, more specifically relying on \gls{sb3}~\cite{stable-baselines3}, and a simulated environment of the Quanser Aero~2 based on MATLAB/Simulink. We also want to show that this policy can be transferred to the real system. In the end, the generated policy is therefore not only able to control the simulated environment based in Simulink, but also the real system.

\begin{figure}
        \begin{subfigure}{0.49\textwidth}
            \centering
            \includegraphics[width=\textwidth]{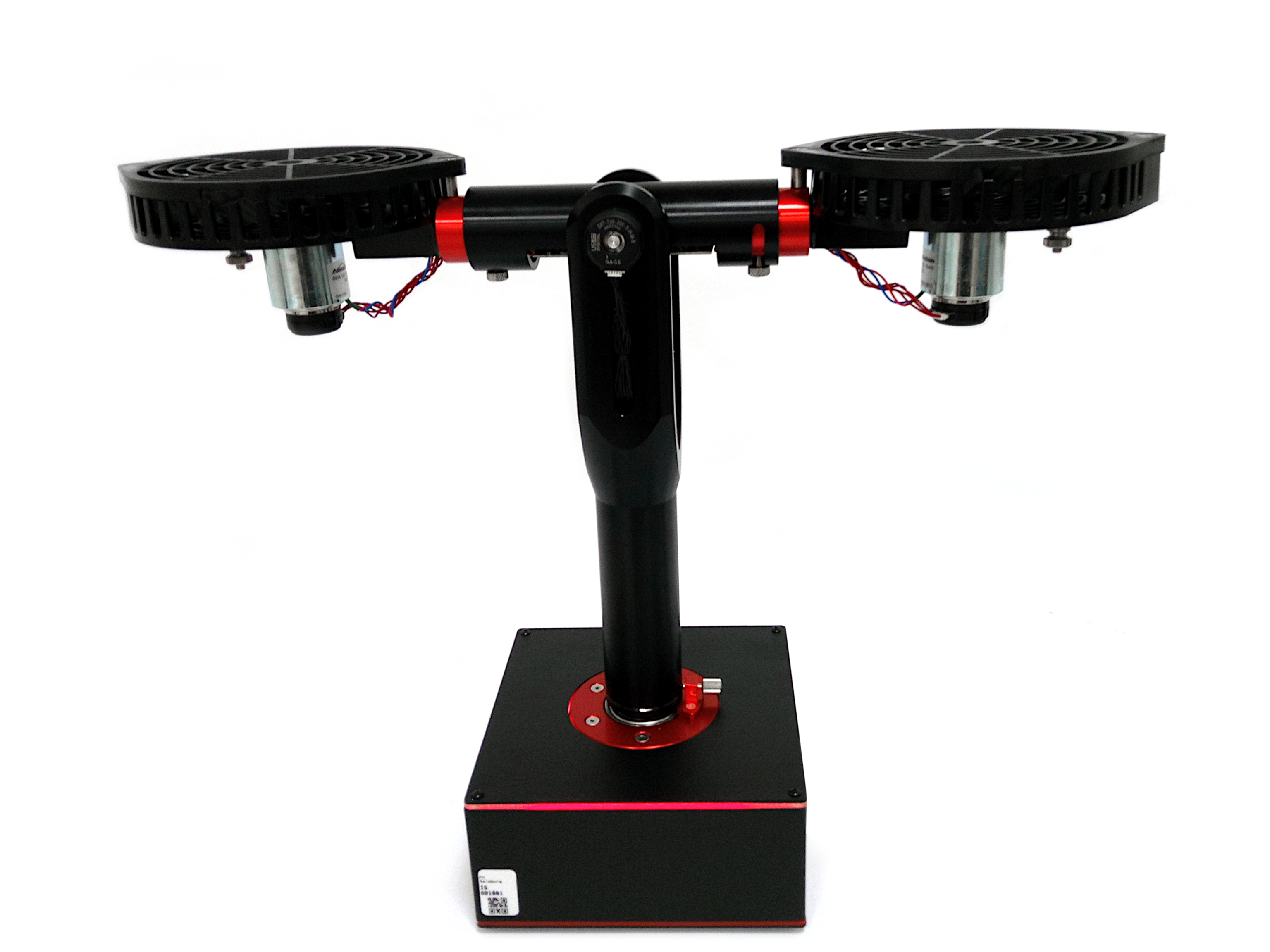}
        \end{subfigure}
        \begin{subfigure}{0.49\textwidth}
            \centering
            \includegraphics[width=\textwidth]{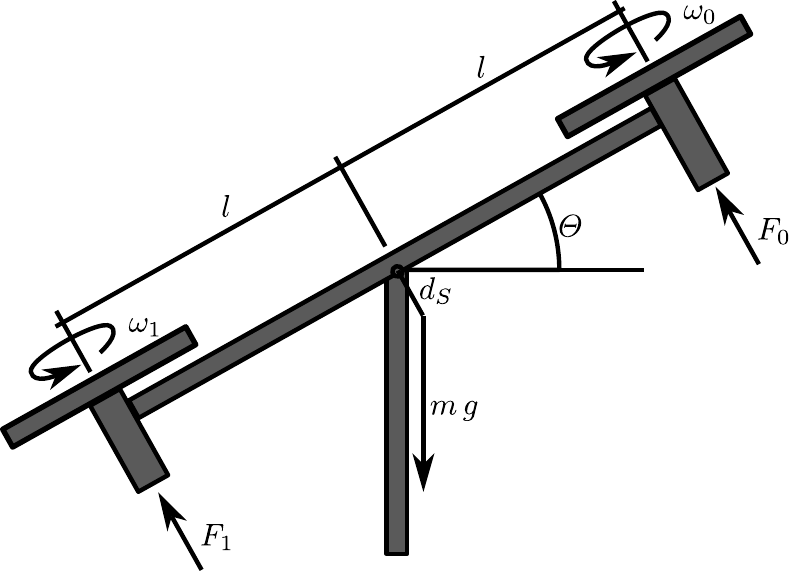}
        \end{subfigure}
        \caption{The Quanser Aero~2~(left) and its schematic representation (right) in a 1-\gls{dof} configuration.}
        \label{fig:aero2}
    \end{figure}

    This specific test bed, the aforementioned Aero~2, is a versatile dual-rotor helicopter and is shown in \cref{fig:aero2}. The system is interesting for \gls{rl} related research since it
    \begin{itemize}
        \item supports multiple inputs and outputs (MIMO),
        \item is a non-linear system,
        \item may be configured in different ways concerning its axes' \gls{dof} and
        \item provides a system model, allowing training in a (Simulink) simulation.
    \end{itemize}

    Using a test bed like this represents industry standards, as simulations and mechatronical models of real systems are available in MATLAB/Simulink. This work aims to expand upon this standard by introducing Python-based \gls{rl}. More specifically, starting out in a 1-\gls{dof} configuration of the Quanser Aero~2, allows us to train a \gls{rl} policy on a simple balancing task, evaluating the feasibility of Python frameworks for \gls{rl} tasks. In future research, learning more complex balancing tasks (with 2-\gls{dof}) can be achieved by simply changing the settings of the real system and its simulation.

    \subsection{Related Work}

    \citeauthor{fandel2018development} use the Quanser Aero in a 2-\gls{dof} configuration to solve a balancing task using \gls{rl} based on MATLAB. This paper shows that, in principle, training \gls{rl} agents using the MATLAB ecosystem in connection with Quanser's devices is feasible~\cite{fandel2018development}. \citeauthor{polzounov2020blue} present a similar approach that substitutes Simulink with a self-developed Python-based driver to simulate the environment. The work presented is based on a Qube - Servo 2 system by Quanser and focuses on training \gls{rl} agents on the real system~\cite{polzounov2020blue}. In \cite{Schaefer24}, the authors compare the performance of fine-tuned \gls{rl} agents to the performance of \gls{mpc}-based controllers in a 1-\gls{dof} balancing task using the Aero~2. The work is purely based on MATLAB and shows that the default configuration of \gls{ppo} agent in MATLAB is insufficient to solve this basic task without fine-tuning. Even tough the complexity of the task was reduced to a minimum, custom parametrization was needed. This is unexpected, as \gls{ppo} is known for its stability.

    \subsection{Main Contribution}
    In essence, we aim to expand upon the approach of \cite{polzounov2020blue}, moving away from MATLAB onto Python when training \gls{rl} agents, while relying Python Gymnasium's interfaces to connect to Simulink environments.
    This allows us to integrate the models created via Simulink into the \gls{rl} ecoystems provided by popular Python framework such as \gls{sb3}, without hand-crafting equations that determine the system's behavior ourselves.

    \subsection{Reinforcement Learning Setting}\label{sec:rl_setting}
    The task that should be solved is to control the Aero~2 beam angle ($\varTheta$) to a specific angle ($r$) by adjusting the voltage applied to its motors ($u$ and $-u$). The action $u$ is updated every 0.1 seconds to achieve this goal. The system's state representation consists of the distance to the desired angle ($\Delta = \varTheta - r$) and the current angular velocity ($\omega=\dot{\varTheta}$). The reward function is defined as the negative absolute distance to the target angle ($-|\Delta|$), motivating the agent to minimize the deviation from the desired orientation. This is the same setting that is used in \cite{Schaefer24} and therefore allows the comparison of the performance of agents shown in the present paper to the results presented \cite{Schaefer24}.

    \section{Approach}
    We present a systematic approach for integrating Simulink models with Python \gls{rl} implementations. The process involves four main steps:

    \begin{enumerate}
        \item \textbf{Generating C-Code from Simulink:} Utilizing Simulink Coder~\cite{simulinkcoder} the control model can be translated to C-Code.
        \item \textbf{Compiling to \gls{dll}:} The generated C-Code is compiled into a \gls{dll} enabling seamless integration.
        \item \textbf{Embedding \gls{dll} in Python:} The \gls{dll} is interfaced with Python using ctypes, exposing the following essential functions and data structures:
        \begin{itemize}
            \item \texttt{MODELNAME\_initialize}: Initializes the control model.
            \item \texttt{MODELNAME\_step}: Performs a single step of the control model.
            \item \texttt{MODELNAME\_terminate}: Terminates the control model.
            \item \texttt{MODELNAME\_U}: Accesses input data.
            \item \texttt{MODELNAME\_Y}: Retrieves output data.
        \end{itemize}
        \item \textbf{Creating Gymnasium Environment:} A custom Gymnasium~\cite{towers_gymnasium_2023} environment is constructed to interface with the Simulink model, implementing the following key methods (see \cref{lst:AeroEnv} in the Appendix for details):
        \begin{itemize}
            \item \texttt{\_\_init\_\_}: Defines the action and observation spaces, and initializes the control model.
            \item \texttt{step}: Executes an action, updates the control model, and returns the new state and reward. To comply with the requirement that the agent has a specified sample time, the update of the model has to be executed multiple times, depending on the configured step-size of the Simulink model.
            \item \texttt{reset}: Resets the environment, reinitializes the control model, and clears input data.
            \item \texttt{render}: Provides optional visualization for monitoring system behavior.
        \end{itemize}
    \end{enumerate}

    Using the created custom Gymnasium environment, a policy can be identified by employing state-of-the-art Python \gls{rl} agent implementations.
    
    \subsection{Deployment}

    To be able to deploy the trained policy to the real system, the Quanser Python API~\cite{quanser_python_api} can be utilized alongside the provided \gls{hil} card. As the pitch $\varTheta$ is needed to define both states, ($\Delta = \varTheta - r$) and ($\omega=\dot{\varTheta}$), the pitch encoder needs to be used to read the aforementioned pitch.
    $\omega$ is determined using a second order low-pass filter, as this first derivative of $\varTheta$ can not be captured directly from the real system. By relying on these two states, the greedy action $u$ can be determined using the identified policy learned on the simulated system. In accordance with this action, the voltages $u$ and $-u$ are applied to the corresponding motors by directly writing to the analog channels of the \gls{hil} card. To comply with the sample time of the agent, a software timer is employed, triggering actions accordingly.

    \section{Discussion}

\begin{figure}
        \centering
        \includegraphics[width=\textwidth]{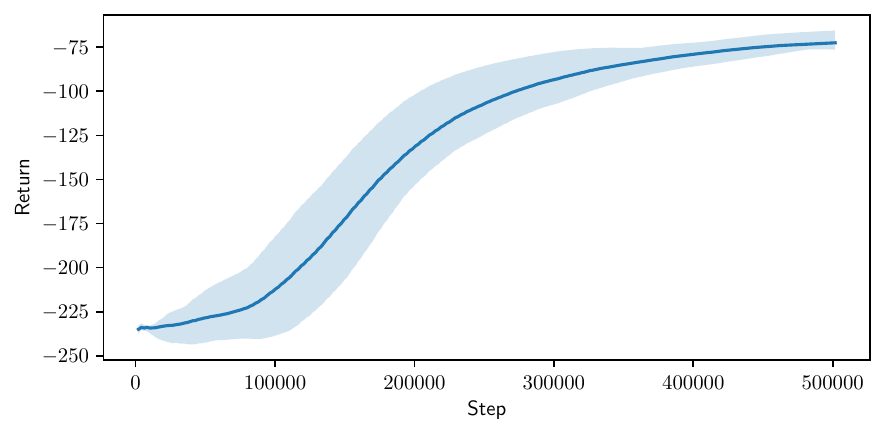}
        \caption{Mean episode return (in dark blue) with minimum and maximum (light blue area) of five training runs of the \gls{sb3} \gls{ppo} agent performed in the simulation of the Quanser Aero~2 system using MATLAB's Simulink in conjunction with Gymnasium.}
        \label{fig:episode_return}
    \end{figure}

\begin{figure}
        \centering
        \includegraphics[width=\textwidth]{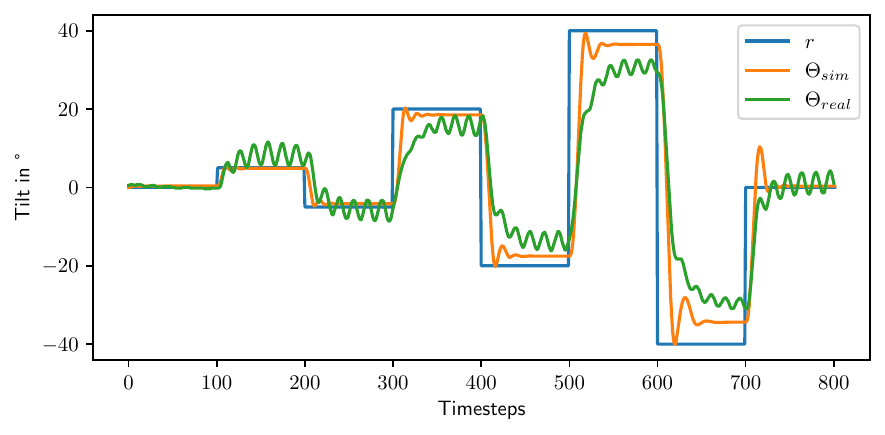}
        \caption{Evaluation runs on the simulation and real system illustrating the behavior of the actual tilt $\varTheta_{\text{sim}}$ and $\varTheta_{\text{real}}$ for simulation and real system, respectively, using a greedy policy and a dynamic target tilt $r$.}
        \label{fig:testrun}
    \end{figure}

    To solve the problem formulated in \cref{sec:rl_setting}, the \gls{sb3}'s \gls{ppo} agent implementation with default parameterization was used. A dynamic tilt function, as illustrated in \cref{fig:testrun}, was used to determine the target pitch $r$. Each training run took 80 seconds, in which every 10 seconds the target pitch changed.
    \Cref{fig:episode_return} illustrates the mean episode return of five independent runs in the simulation and on the real system where each run consisted of 500.000 steps. With the agent's sample time set to 0.1 seconds, each episode takes 800 steps, resulting in 625 episodes per training run. The highest return reached in the run was -64.87, representing a mean average deviation to the target tilt of 4.6°. This is a notable increase in performance compared to prior work from \cite{Schaefer24}, where a maximum reward of -77.93 (average distance to target of 5.6°) was reached.

    \Cref{fig:testrun} shows the evaluation of the agent trained on the simulated model in a real world scenario alongside its performance in the simulation. It is evident that \gls{ppo} is able to solve this task adequately in the simulation environment, and shows promise when applied to the real world system, when keeping in mind that the agent was solely trained on the simulation and therefore had no access to the real world system during training time. 
    Providing an optimal solution to the balancing task, that is solving the problem of steady state deviations as well as oscillations present for both $\Theta_{\text{sim}}$ and $\Theta_{\text{real}}$, is decidedly not in the scope of this work, but is part of our future research. 
    
    \section{Conclusion and Future Work}
    To summarize, we show that training \gls{rl} policies on the basis of a simulated representation of Quanser's Aero~2 using Simulink in combination with Python frameworks such as \gls{sb3} is feasible. Additionally, we have shown that the learned policy is transferable to the real system. Most importantly, the default configuration of \gls{sb3}'s \gls{ppo} implementation outperforms the fine-tuned \gls{ppo} agent based on MATLAB's \gls{rl} Toolbox presented in \cite{Schaefer24}. Note that the default configuration of MATLAB's \gls{ppo} agent was not able to solve the balancing task at all. Being able to rely on the default settings for the \gls{ppo} agent without having to fine-tune hyperparameters significantly decreased development time.
    
    In future work, we want to compare the implementations of \gls{sb3} and MATLAB's \gls{rl} Toolbox \cite{matlabagents} in more detail.
    Additionally, fine-tuning the \gls{sb3} \gls{ppo} agent may lead to increased performance exhibited in the balancing task for both the simulated environment and the real world system. Two additional ways to improve performance for the real system include the expansion of the state space, by adding the actual tilt $\varTheta$, and the training of the \gls{ppo} agent on the real system itself.
    Other planned areas of research include the comparison of the quality of the learned policies balancing capabilities to default controllers such as linear-quadratic regulators or model predictive control. Furthermore, we plan to explore the 2-\gls{dof} configuration of the Quanser Aero~2 system.
    
    \subsubsection*{Acknowledgments}
    Financial support for this study was provided by the Christian Doppler Association (JRC ISIA), the WISS-FH project IAI, the European Interreg Österreich-Bayern project BA0100172 AI4GREEN and the PRISMATICS research project.

    \section*{Appendix}
    \addcontentsline{toc}{section}{Appendix}
    The following listing shows the primary implementation for generating a Gymnasium environment utilizing a DLL generated by Simulink Coder in the specific example of the Quanser Aero~2 system.

        \begin{lstlisting}[caption={Custom Gymnasium Environment for the Aero~2},label={lst:AeroEnv},language=Python]
import gymnasium as gym
import ctypes
import numpy as np
        
class AeroEnv(gym.Env):
  def __init__(self):
    self.action_space = gym.spaces.Box(
      low=-24.0, high=24.0, shape=(1,)
    )
    self.observation_space = gym.spaces.Box(
      low=-np.pi, high=np.pi, shape=(2,)
    )
    self.model = ctypes.CDLL("./aero.dll")
    self.input = get_input(self.model, "aero_U")
    self.output = get_output(self.model, "aero_Y")
    self.model.aero_initialize()

  def step(self, action):
    self.input.v0 = action[0]
    self.input.v1 = action[1]
    for _ in range(5):
      self.model.aero_step()

    return self.state, self.reward, False, False, {}

  def reset(self):
    self.model.aero_terminate()
    self.model.aero_initialize()
    self.input.v0 = 0.0
    self.input.v1 = 0.0
    return self.state, {}

  @property
  def state(self):
    return np.array(
      [self.target_tilt - self.output.pitch, self.output.velocity]
    )

  @property
  def reward(self):
    return -np.abs(self.target_tilt - self.output.pitch)
    \end{lstlisting}

%
%
\printbibliography

\end{document}